\documentclass[journal]{IEEEtran}
\usepackage{bm}
\usepackage{url}
\usepackage{times}
\usepackage{xcolor}
\usepackage{amssymb}
\usepackage{booktabs}
\usepackage{rotating}
\usepackage{footmisc}
\usepackage{setspace}
\usepackage{multirow} 
\usepackage{algorithm} 
\usepackage{algorithmic} 
\usepackage{threeparttable}

\usepackage{makecell}
\usepackage{amsmath,epsfig}
\usepackage{cite}
\usepackage{amsfonts}
\usepackage{graphicx}
\usepackage{stfloats}
\usepackage{subfigure}
\usepackage{listings}
\usepackage{color}
\graphicspath{{figures/}}

\newcommand{\etal}{\textit{et al}. }
\newcommand{\ie}{\textit{i}.\textit{e}. }
\newcommand{\eg}{\textit{e}.\textit{g}. }

\begin{document}

\title{Action Keypoint Network for Efficient Video Recognition}

\author{Xu~Chen, Yahong~Han, Xiaohan~Wang, Yifan~Sun, Yi~Yang 
	\thanks{This work was done during Xu Chen's internship at Baidu Research. \textit{(Corresponding Author:Yahong Han.)}} 
	\thanks{X. Chen and Y. Han are with the Intelligent and Computing College of Tianjin University, Tianjin 300350, China (e-mail: lanncx@gmail.com; yahong@tju.edu.cn).} 
	\thanks{X. Wang and Y, Yang are with the College of Computer Science and Technology, Zhejiang University, Hangzhou 310000, China (e-mail: xiaohan.wang@zju.edu.cn, yangyics@zju.edu.cn).} 
	\thanks{Y. Sun is with the Baidu Research, Beijing 100000, China (e-mail: sunyf15@tsinghua.org.cn).}
}


\maketitle

\begin{abstract}
    Reducing redundancy is crucial for improving the efficiency of video recognition models.
    An effective approach is to select informative content from the holistic video, yielding a popular family of dynamic video recognition methods. 
    However, existing dynamic methods focus on either temporal or spatial selection independently while neglecting a reality that the redundancies are usually spatial and temporal, simultaneously. Moreover, their selected content is usually cropped with fixed shapes (\emph{e.g.}, temporally-cropped frames, spatially-cropped patches), while the realistic distribution of informative content can be much more diverse. With these two insights, this paper proposes to integrate temporal and spatial selection into an \textbf{A}ction \textbf{K}eypoint \textbf{Net}work (AK-Net). 
    From different frames and positions, AK-Net selects some informative points scattered in arbitrary-shaped regions as a set of ``action keypoints'' and then transforms the video recognition into point cloud classification. 
    More concretely, AK-Net has two steps, \emph{i.e.}, the keypoint selection and the point cloud classification. First, it inputs the video into a baseline network and outputs a feature map from an intermediate layer. We view each pixel on this feature map as a spatial-temporal point and select some informative keypoints using self-attention. 
    Second, AK-Net devises a ranking criterion to arrange the keypoints into an ordered 1D sequence. Since the video is represented with a 1D sequence after the specified layer,  
    AK-Net transforms the subsequent layers into a point cloud classification sub-net by compacting the original 2D convolutional kernels into 1D kernels. 
    Consequentially, AK-Net brings two-fold benefits for efficiency: The keypoint selection step collects informative content within arbitrary shapes and increases the efficiency for modeling spatial-temporal dependencies, while the point cloud classification step further reduces the computational cost by compacting the convolutional kernels. 
    Experimental results show that AK-Net can consistently improve the efficiency and performance of baseline methods on several video recognition benchmarks.

\end{abstract}

\IEEEpeerreviewmaketitle

\section{INTRODUCTION} \label{sec:intro}
Video recognition is the foundation of many intelligent video analysis applications in surveillance, entertainment, and security.
In the last few years, the state of the art has been advanced significantly by deep learning \cite{carreira2017quo, feichtenhofer2019slowfast}.
However, the great computational complexity remains a key challenge in realistic video recognition applications. It is because video recognition is based on video data and requires dense spatial-temporal modeling. In response, many recent literature \cite{lin2019tsm, zhu2020faster, wu2019multi, zheng2020dynamic, fan2018watching} challenge the efficiency problem in deep video recognition to promote real-world applications. 

\begin{figure}
	\centering
	\includegraphics[trim=20mm 0mm 30mm 0mm,clip,width=250pt]{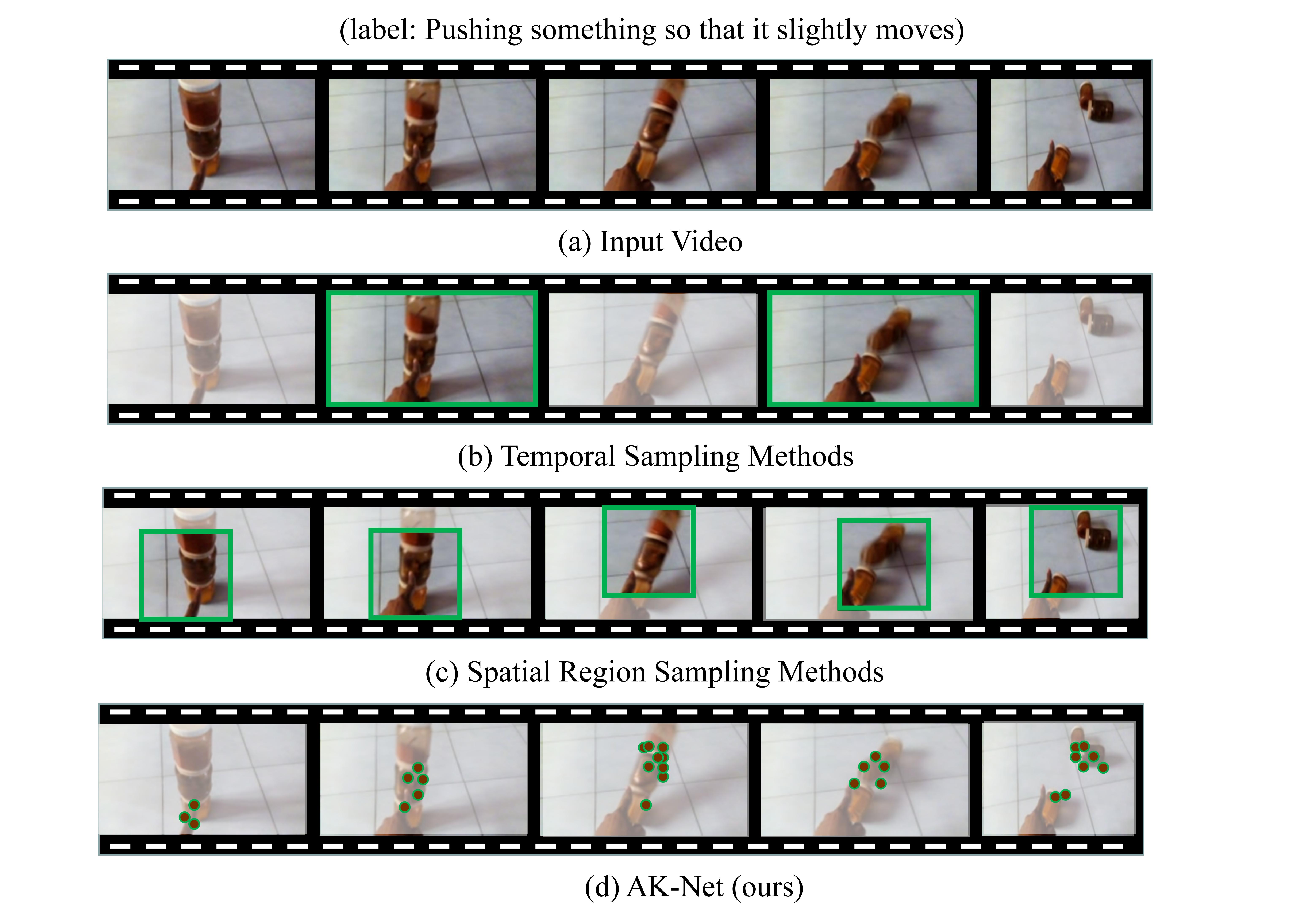}
	\caption{
		\textbf{Illustration of different dynamic video recognition pipelines.}
		This video shows the action of `pushing something so that it slightly moves', and the regions enclosed by green lines are selected by the corresponding models.
		Previous works mainly save computation by ignoring trivial regions with the fixed shape on the temporal or spatial dimension. 
		Our method dives into utilizing finer spatial-temporal points to represent the task-relevant dynamics.
	}
	\label{fig:motivation}
\end{figure}

A popular and effective approach to improve the efficiency is the dynamic video recognition \cite{korbar2019scsampler, wu2019multi, zheng2020dynamic}. Generally, dynamic video recognition selects the informative content from the holistic video and thus reduces the redundancy. Some methods \cite{korbar2019scsampler, wu2019multi} crop the informative frames from the raw video and may be viewed as temporal selection (Fig.~\ref{fig:motivation} (b)). In contrast, \cite{Wang_2021_AdaFocus} crops informative patch from each frame and may be viewed as spatial selection (Fig.~\ref{fig:motivation} (c)). By either temporal or spatial selection, these dynamic methods allocate the computation to partial data instead of equally processing all data volumes. 

We argue that individual temporal selection or spatial selection still has some gap towards accurate and effective selection of informative content. It is because in reality, the redundancies are usually spatial and temporal, simultaneously. \textcolor{black}{For example, when deciding whether there are foul actions in sports videos, we only need to focus on a few areas near the players for key frames, while most content of irrelevant frames can be safely skipped}. Moreover, existing dynamic methods use fixed shapes (bounding boxes, in particular) for selection and lacks flexibility to capture informative content which may distribute in arbitrary-shaped regions. Therefore, it is important to combine both spatial and temporal selection, as well as to improve their flexibility.

Such motivated, this paper proposes a novel Action Keypoint Network (AK-Net) to integrate spatial and temporal selection for efficient video recognition. Instead of selecting some whole frames or whole patches, AK-Net uses ``point'' as the basic unit for extracting informative content, as illustrated in Fig.~\ref{fig:motivation} (d). These points can be selected from any spatial locations within any temporal frames, therefore bringing high flexibility for temporal-spatial integration. 
We note that using point as the basic representation unit shares inspiration from the traditional space-time interest points (STIPs). However, the differences between STIPs and the proposed AK-Net are significant and fundamental. In STIPs, the extracted salient points \cite{bregonzio2009recognising} or cuboids \cite{dollar2005behavior} are obtained via manual designed detectors and are extracted on the raw data. In contrast, AK-Net generalizes such inspiration into deep learning and conducts point-wise selection in the deep feature space. To imply these differences from STIPs, in this paper, we name these informative points as action keypoints, and name the proposed method as Action Keypoint Network (AK-Net). 

During the design of AK-Net, we mainly focus on two crucial questions:

\textbf{1) How to select the action keypoints.}
Current deep neural networks can explicitly learn the localizable representation inherently \cite{zhou2016learning, selvaraju2017grad}, which prompts us to detect salient points from the activated feature map directly.
To be specific, we aggregate the feature map of intermediate layer along the channel dimension to obtain the heatmap.
The derived heatmap indicate the confidence score for each spatial-temporal position.
Then we select top $N$ points as the action keypoints by considering scores across all frames.
However, the selection operation is non-differentiable, only the selected point can be optimized during training.
To alleviate this problem, an auxiliary classification head is introduced for early prediction, which allows back-propagation for every point.
Before sending the intermediate feature to the classification head, it will be reweighted by the heatmap, which encourages the task-related points to get higher scores.
The selected action keypoints are highly representative, we find that our method can surpass the baseline method while only using 30\% points of the whole feature map.

\textbf{2) How to classify actions using point features.}
The keypoint selection breaks the original spatial-temporal structure and makes it difficult to process the keypoints with the original convolutional layers. Specifically, in the original structure, each frame contains fixed number of points and all the frames are stacked in a sequential order. After the keypoint selection, different frames might contain different number of keypoints, and the keypoints in each single frame are not well-aligned. In response to this data structure change, we make two changes to further process these keypoints and to classify actions. First, we re-arrange these keypoints into an ordered sequence of point features using a newly-designed criterion. The ordered sequence has the same data structure of point cloud. Second, we transform the sub-sequential layers of the baseline model into a point cloud classification sub-net by compacting the original 2D convolutional kernels in to their 1D counterparts.  
For example, if the spatial size of a 2D convolutional kernel is $3 \times 3$, the spatial size of its 1D counterpart is $3$. Please note that this compacting operation does not change the topology structure of the original layers. Moreover, it allows using the pre-trained 2D kernels for initialization by summing up the ImageNet \cite{deng2009imagenet} pre-trained kernels along a spatial dimension. 
With these designs, AK-Net implements the video recognition tasks in a point cloud classification manner.

AK-Net has two-fold advantages for efficient action recognition. On the one hand, the keypoint selection step collects informative content within arbitrary shapes and neglects a considerable portion of data. On the other hand, the point cloud classification step compacts the 2D convolutional kernels into 1D kernels and thus further reduces the computational cost. Depending on the position of the intermediate layer for keypoint selection, AK-Net improves the efficiency and yet maintains comparable or even slightly higher accuracy on several video recognition benchmarks. 

Our contributions can be summarized as follows:
\begin{itemize}
    \item We propose a novel framework named AK-Net to simultaneously reduce temporal and spatial redundancy for efficient action recognition. Ak-Net selects some informative feature points (\emph{i.e.}, action keypoints) from the holistic convolutional features to efficiently capture spatial-temporal dependencies. 
    \item Apart from the keypoint selection, AK-Net brings another advantage by transforming the action recognition task into point cloud classification. The selected keypoints are arranged into ordered point cloud and require only 1D convolution processing, which is more efficient than the 2D baseline. 
    \item We show that AK-Net is compatible to many video recognition backbones and consistently improves the recognition efficiency. On Something-Something V1 \& V2 \cite{goyal2017something} and Diving48 \cite{li2018resound}, it reduces the computational cost while maintaining (or slightly increasing) the video recognition accuracy. 
\end{itemize}

The remainder of this paper is organized as follows.
In Section \ref{sec:relatedwork}, we briefly review related works about efficient video recognition.
In Section \ref{sec:method}, we introduce the pipeline of the proposed method in detail.
In Section \ref{sec:experiment}, we provide experimental results including ablation studies, visualization, and comparisons to verify the effectiveness of our method.
Finally, we conclude our work in Section \ref{sec:conclusion}.

\section{RELATED WORK} \label{sec:relatedwork}
\subsection{Video Recognition}
Video recognition is a topic with long history in computer vision community, it aims to predict the corresponding category for a given video.
In the early age, most methods focus on constructing hand-crafted features by considering spatial and temporal correlations to benefit the recognition \cite{wang2013action, bregonzio2009recognising}, and the feature extraction and classification are always separated.
In recent years, video recognition has been greatly developed by deep learning-based paradigm, which integrate the feature extraction and classification into a whole network and optimize all components in an end-to-end manner.
Simonyan and Zisserman \cite{simonyan2014two} proposed a two-stream architecture to utilize RGB and optical flow with two independent 2D Convolutional Neural Network (CNN), and then fused their scores for final prediction.
Instead of extracting appearance and motion information using two 2D network, Tran \etal \cite{tran2015learning} proposed a 3D CNN to jointly model spatial and temporal information.
However, 3D models have more parameters, which incurs more difficulties during optimization with limited data.
To alleviate the training problem of 3D networks, Carreira and Zisserman \cite{carreira2017quo} inflated kernels of 2D networks to 3D, which allowed a better utilization for the solid prior knowledge learned from image data.
Besides, they proposed a large-scale video dataset, named Kinetics, to facilitate the development of deep learning-based action recognition.
Currently, 3D models are prevalent in state-of-the-art video analysis frameworks \cite{ji2019context, wang2021interactive, wang2020symbiotic, chen2021infrared, xu2018sequential, wang2021t2vlad, wu2020learning} due to its solid performance for spatial-temporal modeling.

Since 3D CNN-based models are too expensive, much effort has been devoted for efficient video analysis.
Among them, exploring more compact representation is an important direction.
Some methods reduced the computation by decreasing the complexity of 3D convolutional kernels \cite{carreira2017quo, hara2018can}, such as decomposing the 3D convolution kernel into 2D and 1D \cite{xie2018rethinking, qiu2017learning}, or using wavelet transform to decompose the kernel \cite{chen2020frequency}.
Some methods \cite{tran2019video, luo2019grouped, kumawat2021depthwise} tried to lighten the architecture by introducing group or depth-wise convolution.
Tran \etal \cite{tran2019video} proposed to utilize the depth-wise convolution to replace the 3D convolution of standard bottleneck block.
Luo \etal \cite{luo2019grouped} split the feature map into spatial and temporal groups in parallel, and utilized a gate to learn the weights of each branch.

Another group of methods aim to extract frame-level features based on 2D network, and the temporal correlations can be captured by late fusion \cite{wang2016temporal} or adding temporal convolutions on intermediate layers \cite{lin2019tsm,li2020tea,jiang2019stm,sudhakaran2020gate}.
Lin \etal \cite{lin2019tsm} directly adopted the 2D CNN equipped with the temporal shift operation.
Such implementation allowed the parallel feature of each frame to interact with others by exchange partial channels, which could guide the original 2D convolution to learn spatial-temporal information without additional parameters.
Sudhakaran \etal \cite{sudhakaran2020gate} further combined the idea of \cite{lin2019tsm} and \cite{luo2019grouped}, they designed a gate-shift module that can route features through time adaptively.
Many methods tried another way that introduced the flow information for representation learning.
Li \etal \cite{li2020tea} has further explored the motion excitation module and multiple temporal aggregation module to capture short- and long-range temporal information respectively.
Zhu et al. \cite{zhu2021temporal} used a cross-layer attention mechanism to incorporate multi-layer features for multi-scale spatio-temporal learning. 
In \cite{wang2021tdn}, Wang \etal built a two-level difference modeling paradigm to capture multi-scale temporal information.
Although effective, the frame-by-frame and pixel-by-pixel processing in above models is highly redundant, since many actions only focus on a small part of regions, no matter foreground or background \cite{yue2018compact, zeng2019graph}.

\subsection{Reducing Spatial and Temporal Redundancy}
For videos, considerable redundancy existed in both temporal and spatial level, thus selecting informative frames or image patches is a potential way to reduce total computations.
Previous methods mainly focus on the temporal reduction by adding a network to sample important frames for fine video recognition.
Korbar \etal \cite{korbar2019scsampler} introduced the reinforcement learning paradigm to identify the most salient temporal clips in a long video.
Multi-agent has also been proposed in \cite{wu2019multi} to formulate the frame sampling as multiple parallel Markov decision processes.
More recently, Wu \etal \cite{wu2020dynamic} trained a policy network which contains a long short-term memory to provide the context information, it can interact with the video sequence and decide which frames to use dynamically.
Besides, another group of methods focus on reducing spatial redundancy.
Meng \etal \cite{meng2020ar} employed a policy network to adopt different resolution for each frames.
The devised network can set appropriate resolution according to the input frame at each timestamp.
Habibian \etal \cite{habibian2021skip} proposed to skip background regions by designing a binary gate for each layer to decide whether the regions were important for model prediction.
Wang \etal \cite{Wang_2021_AdaFocus} allocated the major computation to task-relevant regions via a global CNN and a policy network.
This method can be further extended to spatial-temporal patch selection and save more computation.
With the sparse sampling strategy and corresponding conditions, above methods can save substantial computation while maintaining accuracy for fast video recognition.
However, both frame and patch selection are not flexible enough for variant dynamics in videos since the distribution of task-related regions are always diverse.
In this paper, we propose to jointly reduce the spatial-temporal redundancy using point features, which can flexibly capture the motion information at any location.
This idea shares the same spirit of multiple knowledge representation \cite{yang2021multi}, where injecting different sources of knowledge is beneficial to the original model.
To our best knowledge, it has not been exploited for efficient video recognition.

Except the methods mentioned above, our method is most related to the stochastic spatial sampling method \cite{xie2020spatially} for 2D images, point cloud classification \cite{qi2017pointnet, qi2017pointnet++}, and skeleton-based action recognition \cite{shi2020skeleton, bian2021structural}.
The main differences are:
1) We mainly focus on processing videos and we jointly select spatial and temporal points from video features with a confidence guided manner.
2) The generated point features have more dimensions, which are quit different from the 3D point coordinates in point cloud classification.
3) Skeleton-based action recognition aims to predict labels by utilizing the human keypoints detected by pose detectors like OpenPose \cite{cao2017realtime}, while we generate action keypoints on the semantic feature map.

\section{METHODOLOGY}\label{sec:method}

\begin{figure*}[ht]
	\centering
	\includegraphics[trim=0mm 10mm 0mm 15mm,clip,width=480pt]{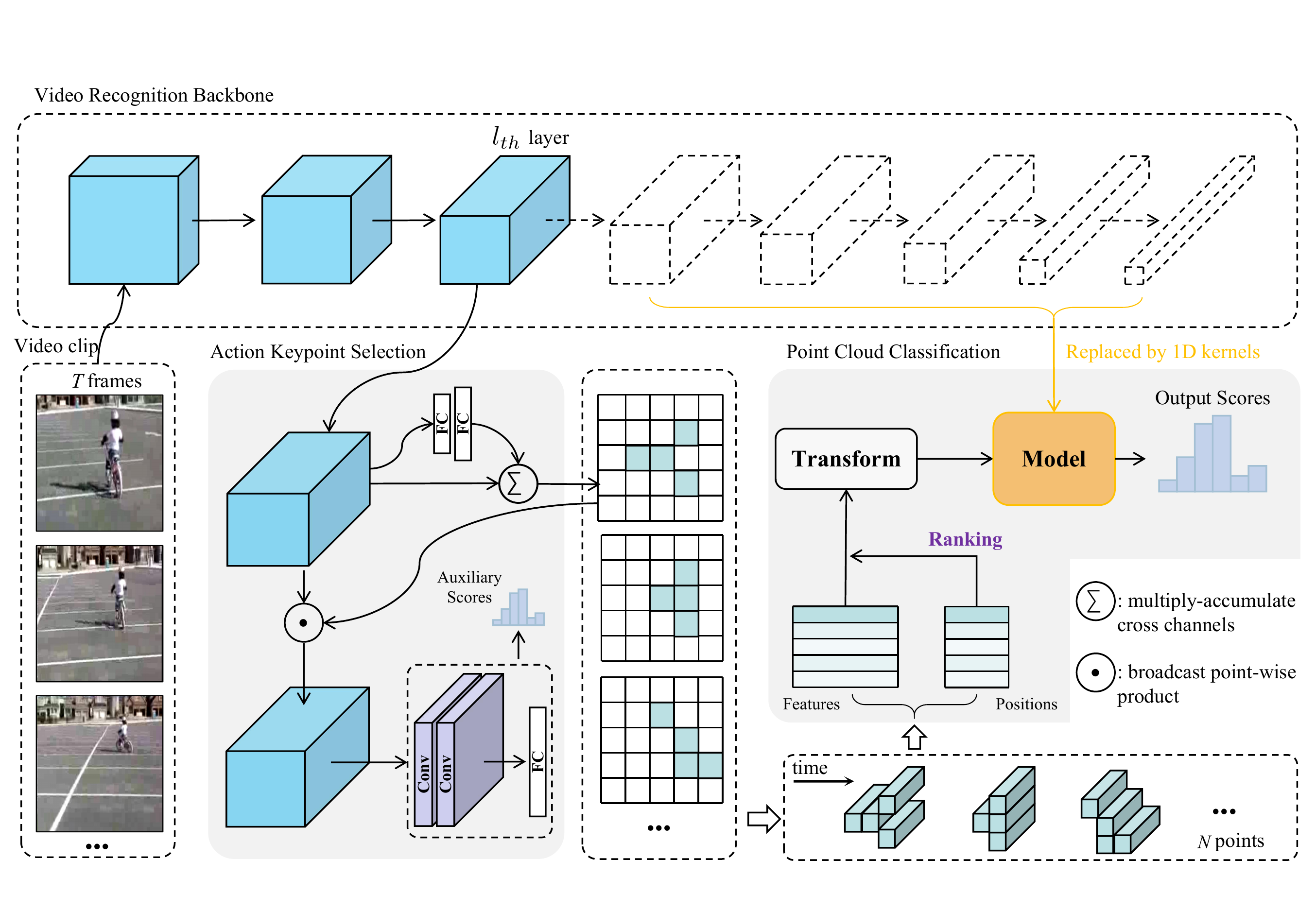}
	\caption{
	Overview of the proposed AK-Net. 
	It is built on the existing video recognition backbone and the whole network is divided into two parts.
	The first part aims to select action keypoints by utilizing the spatial-temporal heatmap.
	An auxiliary classification head is introduced to facilitate the optimization of all point features.
	Then a set of spatial-temporal points are selected via the score of the heatmap. 
	The selected action keypoints will be processed by the second part, which is composed of ranking, transform, and 1D convolution layers.
	}
	\label{fig:framework}
\end{figure*}

The overall pipeline of AK-Net is shown in Figure \ref{fig:framework}.
It consists of two sequential steps---action keypoint selection and point cloud classification.
We build these two parts on existing video recognition backbone (\emph{e.g.}, TSM \cite{lin2019tsm} and TDN \cite{wang2021tdn}) and divide the backbone into two sequential parts corresponding to these two steps. Specifically, we use the bottom layers before a specified intermediate layer as the keypoint extraction sub-net and transform the top layers after the intermediate layer into the point cloud classification sub-net. 


\subsection{Action Keypoint Selection.}
For a basic video recognition backbone with $L$ layers, the mapping function of the whole network can be denoted as $\mathcal{G}=g_1\circ g_2 \cdots \circ g_L$, where $g$ is the mapping function of each layer, $g_i\circ g_{i+1}$ denote the function composition: $g_{i+1}$ after $g_i$.
Given the video $V = [I_1, I_2, ..., I_T]$, where $I_t\in \mathbb{R}^{3\times h\times w}$ is the frame at time $t$ and $T$ is the length of the video sequence, the action category can be predicted by:
\begin{equation}
	\hat{y} = \mathcal{G}(I_1,I_2,\cdots,I_T; g_{\theta_1}, g_{\theta_2}, \cdots, g_{\theta_L}),
\end{equation}
where $g_{\theta_i}$ is the corresponding weight for $g_i$, and $\hat{y}$ is the predicted action label.
In our proposed setting, we divide the entire network into two sequential sub-networks, and the shallow layers are employed to extract the primary video feature:
\begin{align}
    \mathcal{G}&\rightarrow\mathcal{F}\circ\mathcal{P}, \\
	X_l = \mathcal{F}&(I_1,I_2,\cdots,I_T; \Theta_f),
\end{align}
where $\mathcal{F}$ and $\mathcal{P}$ denote the mappings of shallow layers and deeper layers respectively.
$\Theta_f=\{g_{\theta_1},g_{\theta_2},\cdots,g_{\theta_l}\}$ is the weight of $\mathcal{F}$, and $X_l$ with shape $C\times T\times H\times W$ is the intermediate feature map at the $l$-th layer.
Then we consider to generate the heatmap from $X_l$ to guide the point selection.
As mentioned before, deep features can learn localizable representations, hence each channel of $X_l$ can activate informative locations in different aspects.
To aggregate these information, we simply apply a weighted summation on $X_l$ along the channel dimension.
To obtain the weight for each channel $X_{l,c}$, we first reduce its spatial size to 1 using the average pooling:
\begin{equation}
    Z_c = \frac{1}{HW}\sum_{i=1}^{H}\sum_{j=1}^{W}X_{l,c}(i,j),
\end{equation}
where $Z_c$ is the reduced value of channel $c$, and the output for each frame is $Z\in \mathbb{R}^{C}$.
Note that we omit the temporal dimension for convenience since the heatmap generation is implemented for all frames in parallel.
Then, we calculate the channel weight by adopting the squeeze and excitation module proposed in \cite{2018senet} and changing its activation function to \textit{Sofxmax}:
\begin{equation}
	\omega = Softmax(W_2 ReLU(W_1Z)),
\end{equation}
where $\omega\in \mathbb{R}^{C}$ is the obtained channel weight, $W1\in \mathbb{R}^{C\times\frac{C}{r}}$ and $W2\in \mathbb{R}^{\frac{C}{r}\times C}$ are the weights of hidden layers and $r$ is the reduction ratio.
Then, the heatmap is generated via the inner product for each location of $X_l$:
\begin{equation}
	\mathcal{H}(i,j) = \omega^\top X_l(i,j),
\end{equation}
where $\mathcal{H}\in \mathbb{R}^{H\times W}$ is the desired heatmap for each frame.
We put the $\mathcal{H}$ of all frames together and sort their scores.
Then we select the top $N$ points, where $N$ is controlled by a factor $\alpha$, \ie, $N=\alpha THW$.
The obtained point feature $\mathbf{P}\in\mathbb{R}^{N\times C}$ will be regarded as a point cloud and we feed it to the 1D classification sub-net, which is described in the following.

\subsection{Point Cloud Classification}

\begin{figure}[ht]
	\centering
	\includegraphics[trim=0mm 70mm 70mm 30mm,clip,width=230pt]{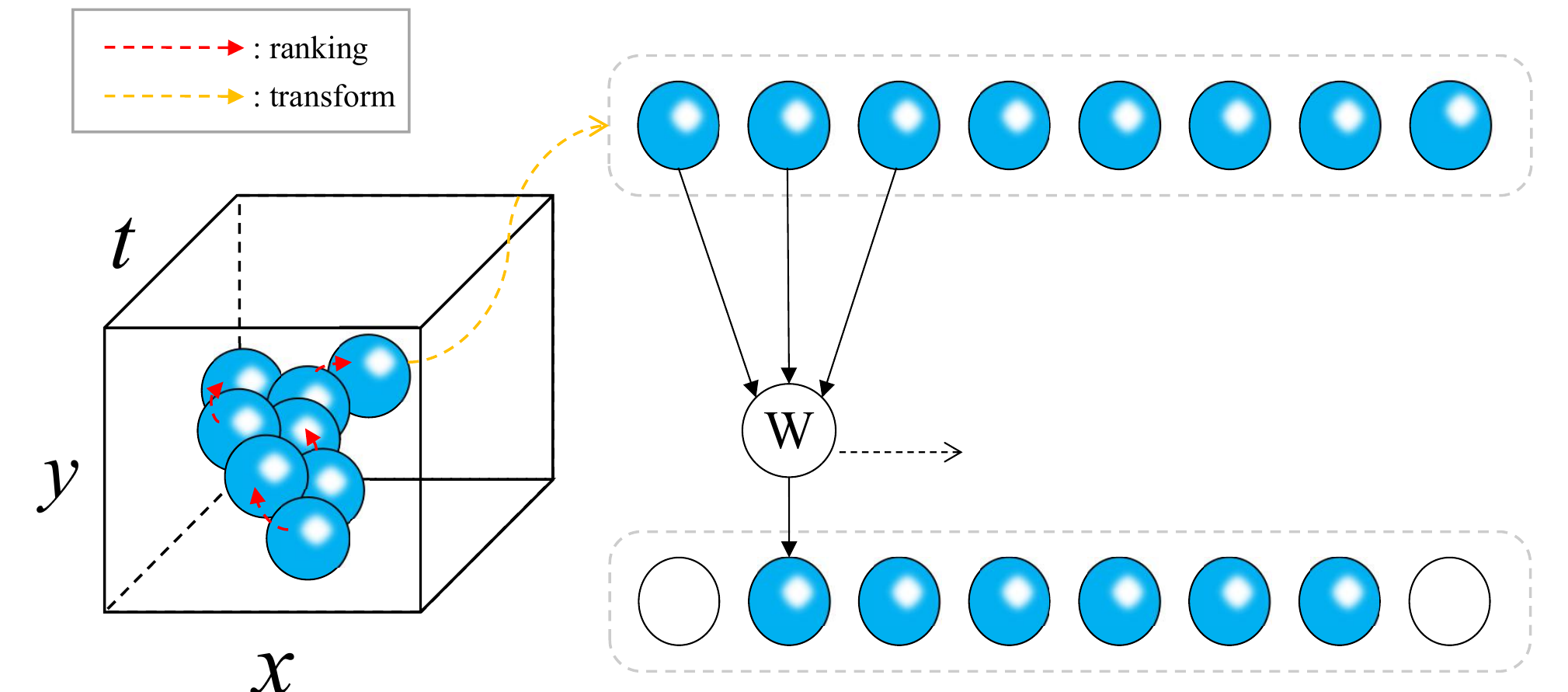}
	\caption{Illustration of the feature extraction for action keypoints.
	The selected points (blue) will first be ranked (red arrows) to reserve the relative local continuity.
	Then we transform (orange arrow) them by jointly considering their coordinates and features.
	Finally, we process these point features using 1D convolution layers.
	}
	\label{fig:rank}
\end{figure}

The procedure of point classification is shown in Figure \ref{fig:rank}.
It consists of ranking, transformation, and point classification.
We will introduce them one by one in this subsection.

\textbf{Ranking.}
For the selected point features with a random order, \ie, $P=[p_1,p_2,\cdots,p_N], p\in\mathbb{R}^C$, we apply a ranking operation $\mathcal{R}(\cdot)$ based on their corresponding coordinates $D=\{d_1,d_2,\cdots,d_N\}$ as:
\begin{equation}
	[p_1,p_2,\cdots,p_N]\overset{\mathcal{R}(D)}{\rightarrow}[s_1,s_2,\cdots,s_N].
\end{equation}
We denote the ranked points as $S\in \mathbb{R}^{N\times C}$.
The purpose of this step is to keep the relative local continuity of input points, which can better fit the local modeling pattern of the subsequent 1D convolution layers (Figure \ref{fig:rank}).
Generally, one original point coordinate $d_i$ is denoted as $(x,y,t)$.
We adjust it according to the weight of 1:1:$\tau$ for $x$-, $y$-, and $t$-axis, respectively.
Specifically, we set $\tau > 1$ so that the spatial continuity will be considered preferentially.
For the choice of $\mathcal{R}$, we just use a simple descending sort based on the summation of weighted coordinates, \ie, $x+y+\tau t$.
We also tried other more complex ranking algorithm like Dijkstra.
However, it takes too much time consumption and does not get a better performance than the simple one.







\textbf{Position and Feature Transformation.}
One key component of general point classification methods is utilizing the position information.
In our method, since the point features lose the original grid structure in the spatial-temporal cube, we have to compensate for it in the later procedure.
A typical way to leverage the position information is adding a learnable embedding, like the transformer-based architectures \cite{vaswani2017attention,dosovitskiy2020image}.
However, in our experiments in Section \ref{sec:ablation}, simply adding the position embedding can not achieve good performance.
In addition, the feature-level transformation is also important, since the feature space of action keypoints has been shifted caused by the selection and ranking operations.
Therefore, we employ a transform network to generate a transform matrix for point features by considering both features and coordinates. 
Following \cite{qi2017pointnet, qi2017pointnet++}, we design a light sub-network to transform the features.
Firstly, for the ranked points with coordinates $D'\in\mathbb{R}^{N\times 3}$, a normalization is conducted as:
\begin{equation}
	\frac{D'-\mu_{D'}}{d_{max}}\rightarrow D',
\end{equation}
where $\mu_D$ is the centroid of all points, and $d_{max}$ is the max Euclidean distance between points and the centroid.
Then, for the ranked points $S$ and their coordinates $D'$, we multiply them by a matrix for all channels:
\begin{align}
	A &= \mathcal{T}(concat(S,D')), \\
	\hat{P} &= concat(S,D')A,s
\end{align}
here $A\in\mathbb{R}^{(C+3)\times(C+3)}$ denotes the transform matrix.
$\mathcal{T}$ denotes the mapping function of the transform network, which is composed of three 1D convolution layers and three fully connected layers.
We output the first $N$ vectors in $\hat{P}$ and denote it as $\tilde{P}$, which represents the transformed feature.

\textbf{Point Classification}
In this paper, we regard the generated action keypoints as a point cloud, which can be processed by the efficient 1D convolution network.
However, general point cloud classification network designed for 3D coordinates is not compatible with high dimensional features.
Thus, we maintain the original structure of the rest part of $\mathcal{G}$, \ie, $\mathcal{P}$, but replace all kernels by their 1D counterparts.
To better leverage the pre-trained weight of $\mathcal{P}$, we transfer them to 1D as the initial weight for point classification network $\mathcal{Q}$:
\begin{equation}
	g_{\theta_{l+1}}, g_{\theta_{l+2}}, \cdots, g_{\theta_{L}} \overset{\mathcal{M}(\cdot)}{\rightarrow}q_{\theta_{1}}, q_{\theta_{2}},\cdots, q_{\theta_{L-l}},
\end{equation}
where $\mathcal{M}$ is the mapping function to transfer the 2D weight to its 1D counterpart.
In this paper, we simply accumulate the weights on one dimension.
For example, for the 2D weights $w_{i,j}$, the processed weight $\hat{w}_{i}$ can be calculated as $\sum_{j=1}^{K}w_{i,j}$, where $K$ is the kernel size.
Then we extract the point feature as:
\begin{equation}
	e_f = \mathcal{Q}(\tilde{P};q_{\theta_{1}}, q_{\theta_{2}},\cdots, q_{\theta_{L-l}}),
\end{equation}
where $e_f\in \mathbb{R}^{N\times C'}$ is the feature after the last convolution layer.
To make a stronger representation for the final prediction, we perform global average pooling on the feature map $X_l$ and combine it with $e_f$:
\begin{equation}
	y_{cls} = f_c(concat(GAP(X_l),e_f)),
\end{equation}
where $f_c$ refers to the prediction function with a single fully connected layer, `GAP' is the global average pooling operation.

\begin{table*}[ht]
	\centering
	\caption{The number of parameters and computational cost of different operations. $k_s$ and $k_t$ is the spatial and temporal size of convolutional kernels, respectively. $k_p$ is the size of the 1D convolution. They are generally set as a small number in networks, \ie, 3 or 1.}
	\label{tab:comp_show}
	\begin{tabular}{ccc}
		\hline
		\textbf{Method}  & \textbf{\#Param}s & \textbf{GFLOPs} \\ \hline
		3D CNN  &  $C_{in}\times C_{out}\times{k_s}^2\times k_t$   &  $C_{in}\times C_{out}\times{k_s}^2\times k_t\times T\times H\times W$   \\
		(2+1)D CNN &   $C_{in}\times C_{out}\times({k_s}^2+ k_t)$  &  ${C_{in}\times C_{out}}\times({k_s}^2+k_t)\times T\times H\times W$   \\
		2D CNN      &   ${C_{in}\times C_{out}}\times{k_s}^2$  &  ${C_{in}\times C_{out}}\times{k_s}^2\times T\times H\times W$   \\ \hline
		Ours    &   $C_{in}\times C_{out}\times k_p$  &   $C_{in}\times C_{out}\times k_p\times N $ \\ \hline
	\end{tabular}
\end{table*}

\subsection{Optimization Objective}
Our point selection operation is deterministic and non-differentiable.
To further benefit the optimization for all points and introduce supervision to guide the heatmap generation, we add an auxiliary head to yield the early prediction.
Firstly, we utilize the heatmap as a soft spatial-temporal attention and apply it to the original feature $X_l$ for each channel $c$:
\begin{equation}
	\tilde{X}_{l,c} = \mathcal{H}\odot X_{l,c},
\end{equation}
where $\odot$ is the element-wise multiplication.
The auxiliary classifiacation head takes $\tilde{X}_l$ as input, and it is composed of two convolution layers for reducing the spatial resolution and a linear layer for predicting action labels.
Besides, it can help the heatmap to focus on action-related regions.
However, the heatmap produces a relative wide response for each activated point, as visualized in \cite{wang2021tdn, sudhakaran2020gate} using class activation mapping \cite{zhou2016learning}, which is not suitable for point selection.
We aim to make the score margin between selected points and others as large as possible.
Therefore, we further introduce a regularization term to facilitate the sparsity of these points.
In the ideal situation, the heatmap should be binary, and the entropy of it should be zero.
Hence, we use the log function to constrain the energy directly, but it is prone to get numerical overflow problems during training.
So we change it to the quadratic function:
\begin{equation}
	\label{eq:reg}
	\mathbf{r}_e = \frac{1}{H\times W}\sum_{i=1,j=1}^{H,W}4\mathcal{H}(i,j)(1-\mathcal{H}(i,j)).
\end{equation}
During the training, this term makes the scores of heatmap converge to 0 or 1, which guarantees a more confident point selection.
The final objective is:
\begin{equation}
	\mathcal{L} = \frac{\mathcal{L}_{cls}(y_{cls}, y)+\mathcal{L}_{aux}(y_{aux},y)}{2}+\mathbf{r}_e,
\end{equation}
where $y_{aux}$ is the auxiliary prediction, $y$ is the ground truth label, $\mathcal{L}_{aux}$ and $\mathcal{L}_{cls}$ are the standard cross-entropy loss applied on auxiliary logits and final logits respectively.

\subsection{Discussion}
We argue that both adaptive selection for frames and image patches can be further improved by directly extracting the spatial-temporal points.
The number of parameters and computational cost of different operations used in general video recognition networks are shown in Table \ref{tab:comp_show}.
We measure the computational cost by Giga Floating-point Operations (GFLOPs), which is a hardware-independent metric and it is highly related to the size of input feature and kernel.
As we can see, traditional 3D or 2D networks for video analysis process each spatial-temporal point equally.
Compared with image processing models, excessive computation can be consumed in this regime with additional dimensions both for data volume and kernel size, which leads previous methods to struggle with making a trade-off between accuracy and speed.
The proposed method directly extracts informative points from the feature map and processes them using 1D convolution layers, which is generally used in point cloud classification task \cite{qi2017pointnet, qi2017pointnet++}.
For this initialization, the number of selected points $N$ is generally far less than the total pixels $T\times H\times W$, and the 1D formed kernel size $k_p$ can further save the computation compared with previous methods using the kernel size of ${k_s}^2\times k_t$ or ${k_s}^2$.

\section{Experiments} \label{sec:experiment}
In this section, we present the experimental results of our AK-Net.
First, we describe the evaluation datasets and implementation details.
Then, we perform ablation studies for the components of AK-Net.
After that, we compare our AK-Net with the existing state-of-the-art methods.
Finally, we show visualization results to further analyze the proposed method.

\begin{table*}[ht]
	\centering
	\caption{Results of different video recognition methods on Something-Something V1 \& V2. 
	`-' means values are not reported, `K400' and `IN' indicate the Kinetics400 and ImageNet datasets.
	We use $\times$ to denote the number of views during inference.}
	\label{tab:sth-sth}
	\begin{tabular}{c|c|c|c|c|c|c} \hline
		\textbf{Method} & \textbf{Backbone} & \textbf{Pre-train} & \textbf{\#Frames} & \textbf{GFLOPs} & \textbf{Sth-Sth V1} & \textbf{Sth-Sth V2} \\ \hline
		I3D \cite{carreira2017quo} & 3D ResNet50 & K400 & 32$\times$2 & 153$\times$2 & 41.6 & - \\
		Non-local I3D \cite{wang2018videos} & 3D ResNet50 & K400 & 32$\times$2 & 168$\times$2 & 44.4 & -\\
		I3D+GCN+NL \cite{wang2018videos} & 3D ResNet50 & K400 & 32$\times$2 & 303$\times$2 & 46.1 & - \\
		ECO(En) \cite{zolfaghari2018eco} & BNInception+3D ResNet18 & K400 & 92 & 267 & 46.4 & - \\ 
		S3D-G \cite{xie2018rethinking} & Inception V1 & IN & 64 & 71 & 48.2 & - \\ \hline
		TSN \cite{wang2016temporal} & ResNet50 & IN & 8 & 33.2 & 19.7 & 27.8 \\
		TRN \cite{zhou2018temporal} & BN-Inception & IN & 8 & 16 & 34.4 & 48.8 \\
		TAM \cite{fan2019more} & bLResNet50 & IN & 16$\times$2 & 47.7$\times$2 & 48.4 & 61.7 \\
		TAN \cite{liu2021tam} & ResNet50 & IN & 8 & 33 & 46.5 & 60.5 \\
		TAN \cite{liu2021tam} & ResNet50 & IN & 16 & 66 & 47.6 & 62.5 \\
		TEINet \cite{liu2020teinet} & ResNet50 & IN & 8 & 33 & 47.4 & 61.3 \\
		TEINet \cite{liu2020teinet} & ResNet50 & IN & 16 & 66 & 49.9 & 62.1 \\
		GST \cite{luo2019grouped} & ResNet50 & IN & 8 & 29 & 47.0 & 61.6 \\
		GST \cite{luo2019grouped} & ResNet50 & IN & 16 & 59 & 48.6 & 62.6 \\
		STM \cite{jiang2019stm} & ResNet50 & IN & 8$\times$30 & 33$\times$30 & 46.4 & 59 \\
		STM \cite{jiang2019stm} & ResNet50 & IN & 16$\times$30 & 67$\times$30 & 50.7 & 64.2 \\
		V4D \cite{zhang2019v4d} & ResNet50 & IN & 8$\times$4 & 167.6 & 50.4 & - \\
		CoConv \cite{chen2021video} & GoogleNet & IN & 16 & 18 & 45.5 & - \\
		RubiksNet \cite{fan2020rubiksnet} & ResNet50 & IN & 8 & 15.8 & 46.4 & 59 \\
		SmallBigNet \cite{li2020smallbignet} & ResNet50 & IN & 8+16 & 157 & 50.4 & 63.3 \\
		TSM \cite{lin2019tsm} & ResNet50 & IN & 8 & 33 & 45.6 & 59.1 \\
		TDN \cite{wang2021tdn} & ResNet50 & IN & 8 & 36 & 52.3 & 64.0 \\	\hline
		AR-Net \cite{meng2020ar}  & ResNet50 & IN & 8 & 41.4 & 18.9 & - \\	
		AdaFuse \cite{meng2020adafuse} & ResNet50 & IN & 8 & 31.5 & 46.8 & 59.8 \\	
		AdaFocus \cite{Wang_2021_AdaFocus} & ResNet50 & IN & 8+12 & 33.7 & 48.1 & 60.7 \\	\hline
		Timsformer \cite{bertasius2021space} & ViT & IN & 8$\times 3$ & 590 & - & 59.5 \\
		Timsformer-HR \cite{bertasius2021space} & ViT & IN & 16$\times 3$ & 5110 & - & 62.5 \\
		Timsformer-L \cite{bertasius2021space} & ViT & IN & 96$\times 3$ & 7140 & - & 62.3 \\
		Swin Transformer \cite{liu2021video} & Swin-B & K400 & 32$\times 3$ & 455$\times 3$ & - & \textbf{69.6} \\
		\hline
		\textbf{Ours}$_{TSM}$ & ResNet50 & IN & 8 & 28 & 47.0  & 60.2\\
		\textbf{Ours}$_{TDN}$ & ResNet50 & IN & 8 & 31 & \textbf{52.5} & 64.3 \\
		\hline
	\end{tabular}
\end{table*}

\subsection{Experimental Setup}
\textbf{Datasets.}
We evaluate our AK-Net on three video datasets---Something-Something v1 \& v2 \cite{goyal2017something}, Diving48 \cite{li2018resound}.
\textbf{Something-Something v1 \& v2} are two large-scale datasets created by crowdsourcing.
The former contains about 100k videos over 174 categories, and the number of videos for training and validation is 86k and 11k, respectively.
The latter contains more videos that consist of around 169k videos for training and 25k videos for validation.
\textbf{Diving48} is a fine-grained video dataset of competitive diving, and it has $\sim$18k trimmed video clips of 48 unambiguous dive sequences standardized by the professional.
This proves to be a challenging task for modern video recognition systems as dives may differ in three stages (takeoff, flight, entry) and thus require modeling of long-term temporal dynamics.

\textbf{Implementation details.}
To build a naive baseline for point-based video recognition, we do not design a fancy data augmentation.
We uniformly sample $T$ frames from the given video during training and testing.
$T$ is 8 in all experiments if not specified.
Following previous works \cite{lin2019tsm, li2020tea, wang2016temporal}, we use \textbf{1-clip and center-crop} during inference for efficiency.
For training, we first scale the spatial size of input frames to $256\times 256$, and then random crop a $224\times 224$ patch.
If not specified, we use ResNet50 \cite{he2016deep} as our backbone and adopt our framework on typical video recognition backbone. \ie, TSM \cite{lin2019tsm} and TDN\cite{wang2021tdn}.
Like previous dynamic video recognition methods \cite{Wang_2021_AdaFocus, meng2020adafuse}, we adopt the two-stage training.
In the first stage, we train the pure backbone without point selection and 1D classification.
The optimizer is SGD and the maximum training epoch is 50.
We set the initial learning rate as 0.01, and it decreases by a factor of 10 at 20 and 40.
In the second stage, we fine-tune all modules using the same training scheme on Something-Something V1 \& V2.
For the smaller Diving48, the Adam optimizer with initial learning rate 5e-4 is used.
The maximum epochs is 45, and the learning rate decreases by a factor of 10 at 20 and 40.

\subsection{Comparison with State-of-the-Arts}
In this section, we focus on comparing our AK-Net with other state-of-the-art methods on Something-Something V1 \& V2 and Diving48.
The point features are generated from the `res4' layer to make a good trade-off between accuracy and inference speed.
The detailed comparison of different location of the separating layer can be seen in Section \ref{sec:ablation}.
The sampling rate $\alpha$ for points is 30\%.
It is expected that feeding more frames can improve performance from previous studies \cite{wang2021tdn, lin2019tsm}.
For a fair comparison with them, we feed the input data with fixed 8 frames that are solely sampled from the RGB stream.
Besides, we keep the backbone as ResNet50, which is generally adopted by efficient models \cite{lin2019tsm, li2020tea, liu2020teinet, wang2021tdn}.

\textit{1) Something-Something V1 \& V2:}
We compare our proposed AK-Net with other temporal modeling methods on Something-Something V1 \& V2 datasets and report the results on Table \ref{tab:sth-sth}.

\textbf{Comparison with 3D CNNs.}
In the first section of Table \ref{tab:sth-sth}, we compare AK-Net with 3D CNNs-based methods.
Generally, video recognition based on 3D CNNs are powerful for capturing spatial-temporal cues and they can achieve high performance while training with large-scale video datasets.
Our AK-Net obtains better performance than them while saving much computation.
Take AK-Net$_{TDN}$ as an example, it outperforms the typical I3D \cite{carreira2017quo} trained with more data (Kinetics-400) by 10.9\% on Something-Something V1 with 9.8$\times$ computation reduction.
For the more efficient (2+1)D variant of I3D---S3D \cite{xie2018rethinking}, AK-Net$_{TDN}$ still obtains 4.3\% accuracy improvement and 2.5$\times$ reduction.

\textbf{Comparison with 2D CNNs.}
AK-Net also shows superiority compared with other efficient methods like 2D CNN-based methods.
As shown in Table \ref{tab:sth-sth}, the proposed AK-Net slightly outperforms its baseline methods, \ie, TSM \cite{lin2019tsm} and TDN \cite{wang2021tdn}, with 13\% and 15\% fewer computation, respectively.
For some extremely lightweight methods, like TRN \cite{zhou2018temporal} and RubiksNet \cite{fan2020rubiksnet}, which can achieve a great reduction of GFLOPs.
However, they failed to achieve good performance compared with other state-of-the-art methods, which indicates that it is difficult to make a trade-off between accuracy and computational cost for current video analysis models.

\textbf{Comparison with dynamic methods.}
In the third section of Table \ref{tab:sth-sth}, we compare three dynamic architectures, which can save computation during inference conditioned on input instances.
Our AK-Net also outperforms them slightly.
In fact, since dynamic methods just allocate a subset of the whole network for each instance, it is difficult to achieve higher performance compared with the complete architecture.
For example, AR-Net \cite{meng2020ar} only get 18.9\% accuracy on the Something-Something V1 dataset, which is even lower than the simply 2D network \cite{wang2016temporal}.
AdaFocus \cite{Wang_2021_AdaFocus} achieves a good trade-off between accuracy and computational cost.
However, to build a strong baseline, it integrates the additional features extracted from MobileNetV2 \cite{sandler2018mobilenetv2} and samples more frames.
Our AK-Net can consistently improve the baseline without changing the original settings (1.1\% and 0.3\% for TSM and TDN on Something-Something V2, respectively).
Moreover, since our method does not employ an additional network, it can be more efficient and general.

\textbf{Comparison with Transformer-based methods.}
We also list the results of current transformer-based methods \cite{bertasius2021space, liu2021video}.
Although they can achieve excellent performances on different vision tasks, it is still too heavy even compared with 3D CNNs.
As shown in Table \ref{tab:sth-sth}, the Video Swin Transformer \cite{liu2021video} can get the highest accuracy on Something-Something V2 dataset, but it requires 44$\times$ computation than our AK-Net$_{TDN}$.
Besides, transformer-based architectures are more data-hungry which needs to be pre-trained on the data with larger scale, like ImageNet21K and Howto100M.

\begin{table}[ht]
    \centering
	\caption{Comparisons with different video recognition methods on Diving48 dataset. `IN', `IN-21K', `K400' denote the ImagNet1K, ImageNet21K and Kinetics-400 respectively, and 'RN101' denotes the ResNet with 101 layers.}
	\label{tab:diving48}
    \begin{tabular}{cccc} \hline
        Method              & Pretrain  & Frames & Accuracy  \\ \hline
        TimeSformer \cite{bertasius2021space}         & IN-21K    & 8$\times$3    & 75        \\ 
        TimeSformer-HR \cite{bertasius2021space}      & IN-21K    & 16$\times$3  & 78        \\ 
        TimeSformer-L \cite{bertasius2021space}       & IN-21K    & 96$\times$3   & 81        \\ 
        I3D  \cite{carreira2017quo} & K400      & 8$\times$1    & 48.3      \\
        ST-S3D  \cite{xie2018rethinking} & K400      & 8$\times$1    & 50.6      \\
        SlowFast,RN101 \cite{feichtenhofer2019slowfast}& K400      & 16$\times$3   & 77.6      \\
        TSN \cite{wang2016temporal} & IN        & 3$\times$1      & 52.5      \\
        GST \cite{luo2019grouped}              & IN        & 8$\times$1    & 78.9      \\
        TSM \cite{lin2019tsm} & IN        & 8$\times$1    & 76.9      \\
        TDN \cite{wang2021tdn} & IN        & 8$\times$1    & 80.5      \\ \hline
        Ours$_{TSM}$          & IN        & 8$\times$1    & 79.3($\uparrow 2.4$) \\
        Ours$_{TDN}$        & IN        & 8$\times$1    & \textbf{81.7}($\uparrow 1.2$) \\ \hline
    \end{tabular}
\end{table}

\begin{figure*}
    \centering
    \includegraphics[trim=30mm 0mm 30mm 20mm,clip,width=480pt]{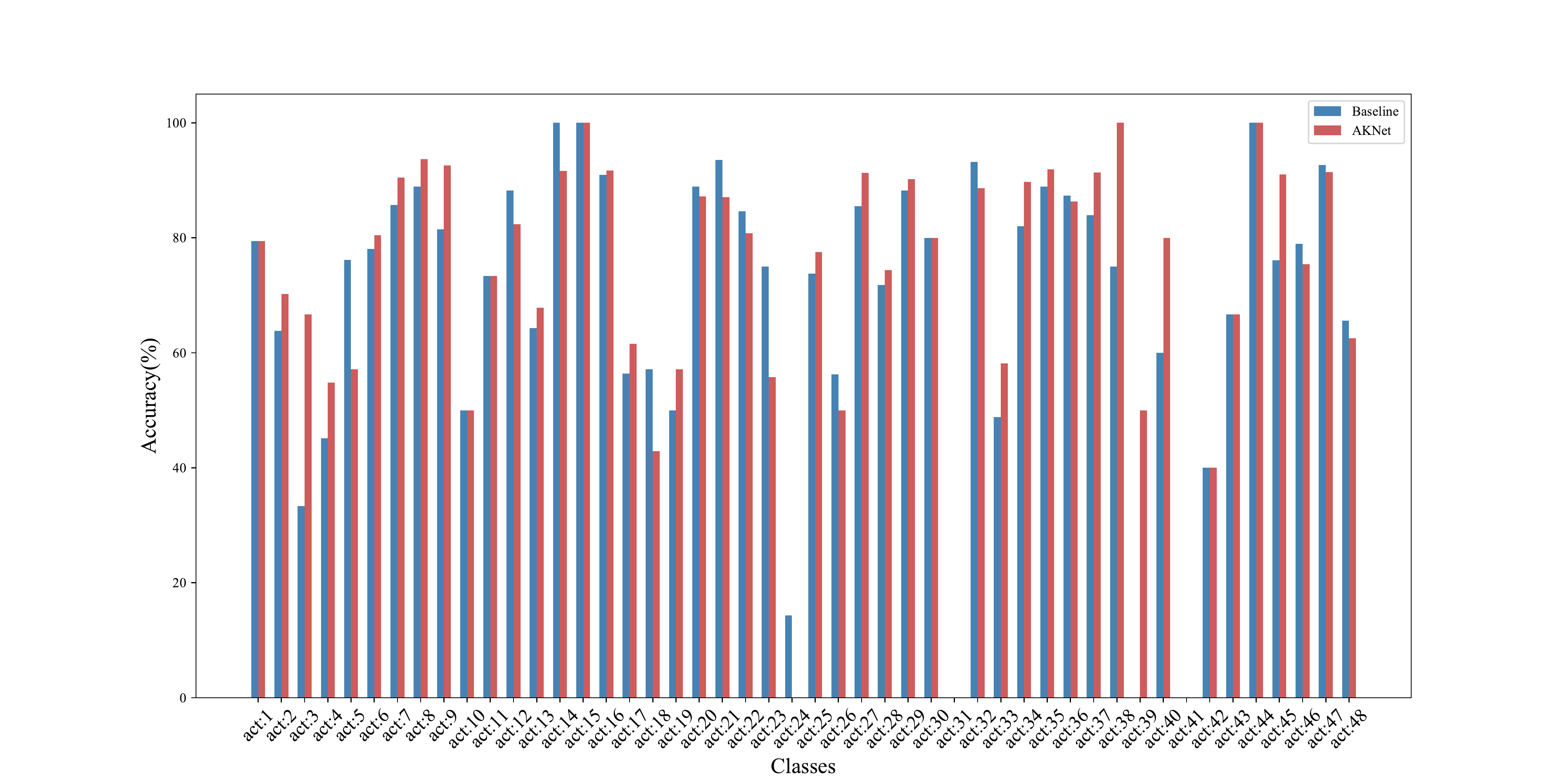}
    \caption{Per class accuracy on Diving48 dataset.
    Because the original action labels annotated by terminologies are not easy to understand, we use the template `act:\#' to replace them for simplicity.}
    \label{fig:per_cls_acc}
\end{figure*}

\textit{2) Diving48}:
We further evaluate our AK-Net on the Diving48 dataset.
Due to the issue discovered in the previous version, we use the corrected annotation of the official updated version.
Results of other methods in Table \ref{tab:diving48} are reported by Bertasius \etal \cite{bertasius2021space}.
As shown in Table \ref{tab:diving48}, AK-Net with TSM and TDN surpass the original methods by 2.4\% and 1.2\% respectively, and it further reduces the computational cost by 15\% and 13\%.
It is consistent with the results in the Something-Something dataset.
Besides, with only given 8 frames as input, AK-Net outperforms current state-of-art methods, even compared with the powerful transformer-based method which trained with more frames and more solid initial weight.
We show the per-class accuracy in Figure \ref{fig:per_cls_acc} for a better comparison with the baseline model.
As we can see, AK-Net surpasses the baseline for most action categories.
Since the action categories in Diving48 are fine-grained, it verifies the strong temporal modeling ability of our proposed method.

\begin{table}[ht]
	\centering
	\caption{Ablations for different components of AK-Net on Something-Something V2.}
	\label{tab:ablation_component}
	\begin{tabular}{cccccc}
		\hline
		w/ $\mathcal{Q}$   & w/ $\mathcal{R}$ & w/ $\mathbf{r}_e$ & w/ $\mathcal{T}$ & concat & Acc(\%) \\ \hline
		\checkmark &    &    &  & & 33.1    \\
		\checkmark & \checkmark  & & & & 45.8    \\
		\checkmark & \checkmark  & \checkmark & & & 46    \\ 
		\checkmark & \checkmark  &  & \checkmark & & 46.9    \\ 
		\checkmark & \checkmark  &  & \checkmark & \checkmark & 47.1    \\ 
		\checkmark & \checkmark  & \checkmark & \checkmark  & \checkmark & \textbf{47.3}    \\ \hline
	\end{tabular}
\end{table}

\begin{figure}[ht]
	\centering
	\includegraphics[trim=10mm 0mm 10mm 5mm,clip,width=250pt]{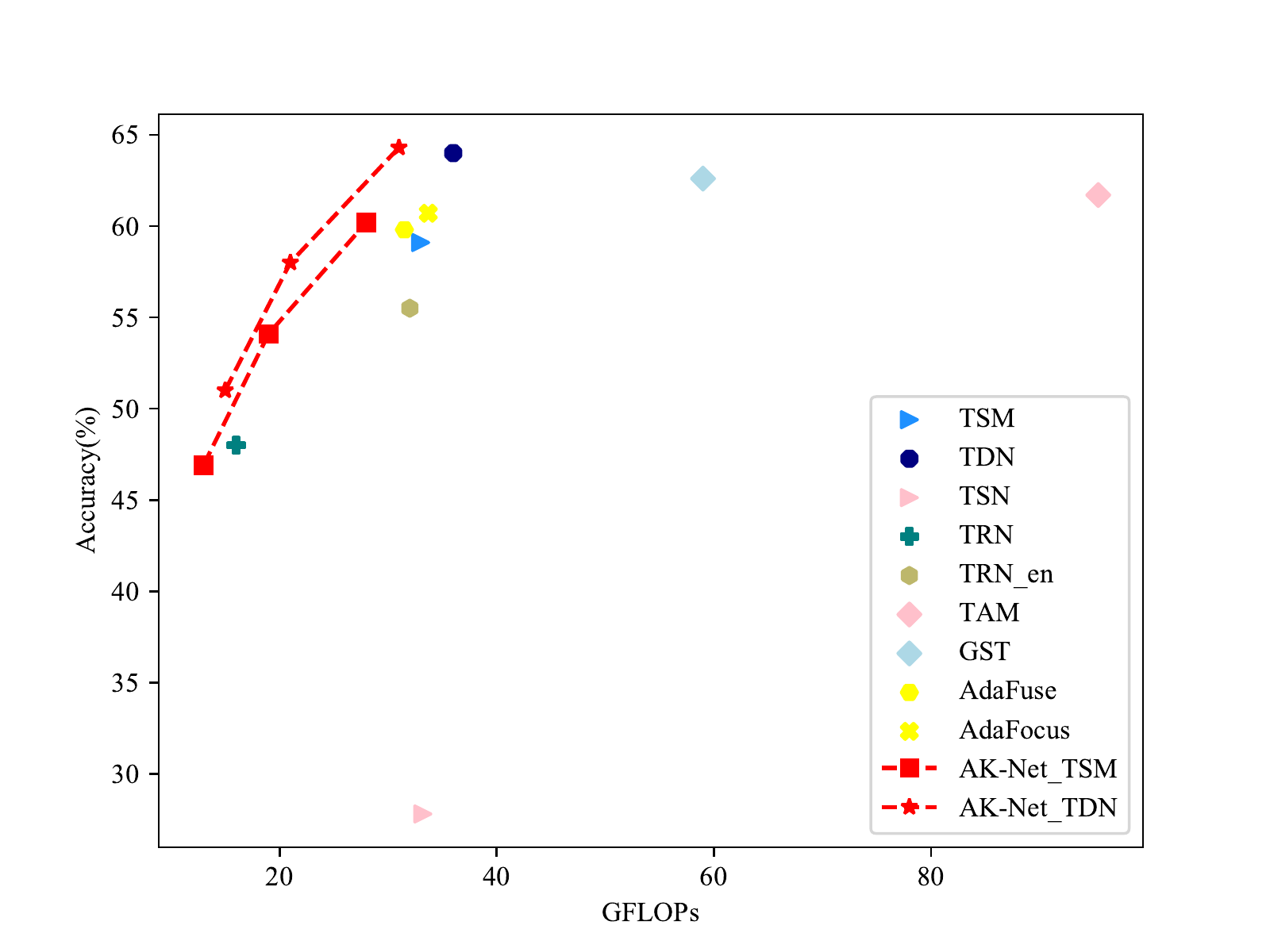}
	\caption{
	AK-Net can achieve different trade-offs (between speed and accuracy) by adjusting the position of separating layer. Compared with different state-of-the-art methods, AK-Net usually achieves higher accuracy with lower computational consumption (GFLOPs), by choosing an appropriate trade-off.} 
	\label{fig:layers}
\end{figure}

\subsection{Ablation Study} \label{sec:ablation}
We perform in-depth ablation studies on Something-Something v2 dataset.
For the following evaluations, we use the inference scheme of one clip and center crop, and Top1 accuracy is reported as the main measurement.

\textbf{Different components.}
We discuss each component proposed in our AK-Net, including the 1D convolution network $\mathcal{Q}$, the ranking operation $\mathcal{R}$, the regularization term $\mathbf{r}_e$, feature transform network $\mathcal{T}$, and the concatenation of intermediate features $X_l$.
Firstly, we solely employ the 1D convolution network to classify point features selected from the output of stage `res2', and it yields a poor performance (33\%).
This implies that only utilizing original features can not recognize the pattern of dynamics well, since the regular grid structure has been lost.
After adding the ranking operation, the performance can be improved with a significant margin (12.7\%). 
The main goal of the ranking function is reserving the local continuity of spatial-temporal points as much as possible.
The energy-based term is proposed to regularize the heatmap to facilitate point selection.
During training, it encourages the points with higher scores to be dominant.
This term only improves the performance slightly (0.2\%).
We guess that the inherent optimization of classification can also achieve the same excitation implicitly.
Besides, we perform global average pooling on the feature map before point selection and add it to the final output features.
By doing this, the discarded points can also take part in the final classification, which complements the low-level semantics and background information.
It can improve the performance by 0.2\%.
Ultimately, the aforementioned components can work jointly and obtain totally 14.2\% improvement compared with the naive pipeline for point classification.

\begin{table}[ht]
    \centering
    \caption{Comparison of different methods for utilizing position information.}
    \label{tab:position}
    \begin{tabular}{cc} \hline
        Method & Accuracy(\%) \\ \hline
        Baseline & 45.3 \\
        +Learnable Embedding & 46.1 \\
        +Feature Transform & 46.9 \\ \hline
    \end{tabular}
\end{table}

\textbf{Position Embedding and Transform.}
To utilize the position information of selected points, we adopt a transform network to consider both features and coordinates.
Here we compare it with another method using the learnable embedding, which is generally used in transformer-based architectures \cite{bertasius2021space, liu2021video}.
We use the AK-Net with ranking operation as the baseline method, and the heatmap is generated from the output of `res2' layer.
As shown in Table \ref{tab:position}, the learnable embedding can improve the baseline by 0.8\%, while the transform network can get a 1.6\% accuracy gain.

\begin{figure*}
    \centering
    \includegraphics[trim=0mm 0mm 0mm 5mm,clip,width=500pt]{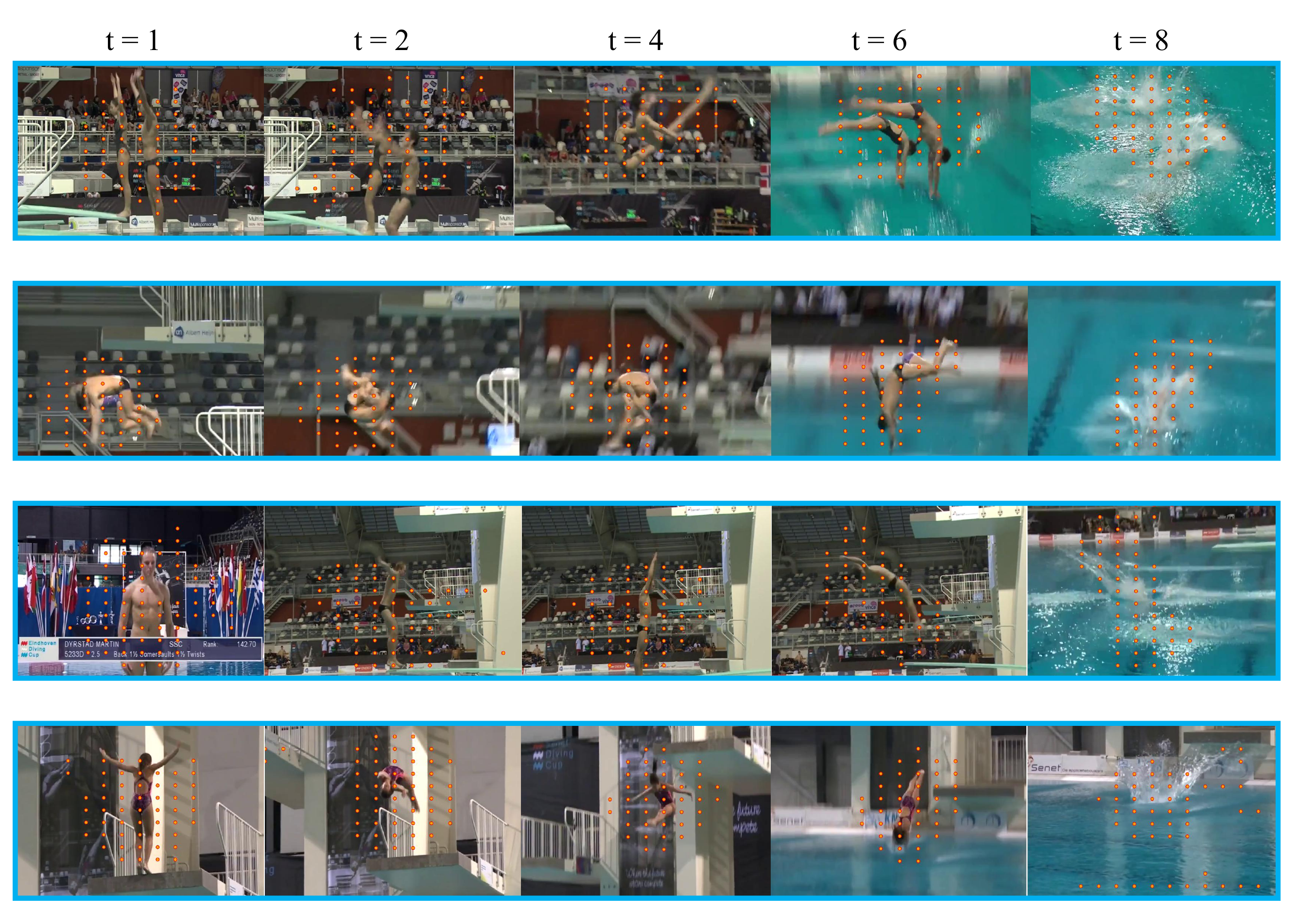}
    \caption{
    Visualization of the estimated action keypoints.
    The keypoint selection is conducted on the `res4' layers, which has a relatively coarse spatial resolution ($14\times 14$ size). Therefore, when they are  mapped onto the raw image ($640\times 480$) for visualization, they look relatively sparse
    (Zoom in for details).
    }
    \label{fig:vis_point}
\end{figure*}

\textbf{Position of the separating layer on video recognition backbone.}
We further investigate the impact of the separating layer position.
For a general residual network, there are five stages, including `conv1', `res2', `res3', `res4', `res5'.
We simply generate point features from the output of these separating layer $l$ except the first and last layers.
As shown in Figure \ref{fig:layers}, for two variants of AK-Net, they achieve higher efficiency with shallower layer, but the accuracy will be relatively dropped.
Our AK-Net can achieve different trade-offs between accuracy and GFLOPs by adjusting the location of separating layer, which meets the computational needs of diverse real-world applications.
Compared with the state-of-the-art methods, AK-Net usually achieves higher accuracy with lower computational consumption (GFLOPs), by choosing an appropriate trade-off. 
For example, when the priority is speed, we use res2 as the separating layer, and the corresponding AK-Net$_{TDN}$ (res2) surpasses the most competitive TRN \emph{w.r.t} both speed and accuracy. 
In contrast, when the priority is accuracy, we use res4 as the separating layer, and the corresponding AK-Net$_{TDN}$ (res4) is faster and more accurate than the most competitive TDN. 

\textbf{Number of action keypoints.}
As shown in Table \ref{tab:point_rate}, we discuss the number of selected points by varying the point sampling ratio $\alpha$.
We employ the AK-Net without ranking and transformation as the baseline.
As the sampling rate grows, the performance of AK-Net first increases and reaches the maximum 48.58\%, then it drops to 44.3\%.
Since we do not perform the same dense modeling for each point and the selected points are highly informative, AK-Net can achieve the highest accuracy while using half of data.
Meanwhile, the baseline needs more points to achieve higher accuracy, which shows a low efficiency.
Our method is superior to the baseline method, with only using $10\%$ points, it even surpasses the baseline that uses all point.

\begin{table}[ht]
    \centering
    \caption{Comparison of different numbers of space-time points.
    The $\alpha$ is the point sampling rate.
    We employ the backbone without ranking and transform operations as `Baseline'.
    }
    \label{tab:point_rate}
	\begin{tabular}{cccccc} \hline
		$\alpha$ & 10\%   & 30\%   & 50\%    & 70\%   & 100\%  \\ \hline
		Baseline & 32.8    & 34.5 & 34.9  & 34.7 & 36.2   \\
		Ours       & 43.2 & 46.9 & \textbf{48.58} & 47.5 & 44.3 \\ \hline
	\end{tabular}
\end{table}

\subsection{Visualization}
We visualize the selected keypoints by mapping them onto the raw videos in Fig \ref{fig:vis_point}.
Here we uniformly sample 8 frames from Diving48 and show the frames at timestamps 1, 2, 4, 6, 8 for simplicity.
Similar to some attention-based methods, our method can detect salient points for each instance during inference.
It is notable that every frame contains a different number of points, which depends on the hardship of input instance.
For some frames with little temporal changes and easy background, \eg, the first frame of the second row, the model can generate fewer points to modeling the dynamics automatically and vice versa.

\section{Conclusion}\label{sec:conclusion}

This paper proposes an Action Keypoint Network (AK-Net) for efficient video recognition. AK-Net  selects informative content from the video features using point-wise granularity. This point-wise selection is capable to remove temporal and spatial redundancy simultaneously and is very flexible. Apart from reducing the redundancy, another benefit towards higher efficiency is using point cloud classification for video recognition. Specifically, since the keypoints are arranged in 1D sequence, AK-Net processes them with 1D convolution and is more efficient than the 2D convolution baseline. In a word, both the keypoint selection and the point cloud classification enhance the video recognition efficiency. Experimental results show that AK-Net consistently improves the recognition efficiency over a battery of video recognition baselines while maintaining comparable recognition accuracy. 

\bibliographystyle{IEEEtran}
\bibliography{IEEEabrv,paper}

\end{document}